\documentclass{article}

\usepackage{microtype}
\usepackage{graphicx}
\usepackage{subfigure}
\usepackage{booktabs} 
\usepackage{xurl}

\usepackage{hyperref}

\usepackage[accepted]{icml2025}

\usepackage{amsmath}
\usepackage{amssymb}
\usepackage{mathtools}
\usepackage{amsthm}

\usepackage[capitalize,noabbrev]{cleveref}

\theoremstyle{plain}

\theoremstyle{definition}

\theoremstyle{remark}

\usepackage[textsize=tiny]{todonotes}

\icmltitlerunning{Building Damage Detection using Satellite Images and Patch-Based
Transformer Methods}

\begin{document}

\twocolumn[
\icmltitle{Building Damage Detection using Satellite Images and Patch-Based Transformer Methods}



\icmlsetsymbol{equal}{*}

\begin{icmlauthorlist}
\icmlauthor{Smriti Siva}{yyy}
\icmlauthor{Jan Cross-Zamirski}{comp}
\end{icmlauthorlist}

\icmlaffiliation{yyy}{Lakeside School}
\icmlaffiliation{comp}{London, United Kingdom}

\icmlkeywords{Machine Learning, ICML}

\vskip 0.3in
]




\begin{abstract}

Rapid building damage assessment is critical for post-disaster response. Damage classification models built on satellite imagery provide a scalable means of obtaining situational awareness. However, label noise and severe class imbalance in satellite data create major challenges. The xBD dataset offers a standardized benchmark for building-level damage across diverse geographic regions. In this study, we evaluate Vision Transformer (ViT) model performance on the xBD dataset, specifically investigating how these models distinguish between types of structural damage when training on noisy, imbalanced data.

In this study, we specifically evaluate DINOv2-small and DeiT for multi-class damage classification. We propose a targeted patch-based pre-processing pipeline to isolate structural features and minimize background noise in training. We adopt a frozen-head fine-tuning strategy to keep computational requirements manageable. Model performance is evaluated through accuracy, precision, recall, and macro-averaged F1 scores. We show that small ViT architectures with our novel training method achieves competitive macro-averaged F1 relative to prior CNN baselines for disaster classification.

\end{abstract}

\printAffiliationsAndNotice{}

\section{Introduction}
\label{Introduction}

Rapid and reliable assessment of building damage after natural disasters is critical in order to get help to the right places quickly. Satellite imagery makes it possible to monitor large areas without needing to get people on the ground. But training reliable damage classifiers is still difficult because the data is noisy, labels are imperfect, and there is a strong skew towards undamaged buildings \citep{fewshot, imbalance, xbd_paper, chen2020}.

The xBD dataset \cite{xbd_paper} was created to address this challenge. It offers pre- and post-disaster satellite images and building-level damage annotations across multiple types of disasters and geographic locations. This makes it a good source for machine learning models for classifying building damage. But this dataset presents some challenges - most images are labeled \textit{``no-damage''} and the differences between minor, major, and destroyed classes is subtle.

Recent work has primarily relied on convolutional neural networks (CNNs) such as ResNet \cite{xbd_paper} and few shot methods \cite{fewshot}. At the same time, Vision Transformers \cite{ViT} such as DINOv2 \cite{DINOV2_paper} and DeiT \cite{DeiT} have shown strong performance on a wide range of vision benchmarks. However, there is still limited empirical evidence on how transformer-based architectures perform on highly-imbalanced datasets \cite{imbalance}.

In this study we make the following contributions: 
\begin{enumerate}
    \item We design a targeted patch-based pre-processing pipeline for the data to remove empty background content from images in training.
    \item We compare fine-tuning strategies for DINOv2-small and DeiT models in constrained compute environments such as Google Colab Pro.
    \item We evaluate the performance of these models across accuracy, precision, recall and F1 metrics.
\end{enumerate}

\section{Background}

\subsection{Related work/Literature Review}

\subsubsection{xBD dataset}

The xBD dataset was introduced as a large-scale benchmark to assess building damage using high-resolution satellite images \cite{xbd_paper}. This covers multiple types of disasters such as earthquakes, hurricanes, wildfires, etc. across diverse geographies. The dataset pairs pre-event and post-event images with building-level polygon annotations and categorical damage labels. In their baseline work, Gupta et al. evaluated CNN models such as ResNet on xBD and reported relatively modest macro-averaged F1 scores. This was due to the strong class imbalance and the subtle visual differences between the damage labels. 

Later work has explored more data-efficient and robust approaches on xBD. This includes few-shot methods \cite{fewshot} for post-earthquake damage detection which improved macro-averaged F1 over the original CNN baselines. These studies collectively make the xBD dataset attractive for benchmarking. At the same time, the dataset is challenging to work with due to the skew towards the \textit{``no-damage''} class, motivating us to explore transformer-based vision models such as DeIT and DINOv2.

Apart from these model-centric improvements, many studies have focused on the imbalance challenge present in the xBD dataset. One example is the two-step architecture proposed by \cite{imbalance}, which explicitly targets the extreme skew in building-damage labels. Their approach introduces a ``normality-imposed data-subset generation" procedure to construct more statistically balanced training samples. By using this preprocessed subset rather than the originally skewed xBD dataset, the model avoids collapsing into the dominant ``no damage"  class and achieves better performance in terms of weighted F1-score. 

This suggests how carefully designed data pre-processing strategies, rather than model architectures alone, can mitigate data imbalance issues and improve the reliability of damage-assessment models. This perspective inspired us to attempt a novel patch-based pre-processing method before training on the xBD data in our research.

\subsection{Vision Transformers}

    The transformer was first introduced in 2016, which proposed the use of attention weights over a convolutional architecture for natural language processing \cite{attention}. The attention mechanism contextually applied weights by computing the similarity scores between tokens, as opposed to the fixed weights in the convolutional method
    \cite{attention}. For each token in an input there is a query, key and value. Similarity scores are gathered through dot-product calculation between the query and key. These scores are used as weights for context-aware outputs.

    \[ \text{Attention}(Q, K, V) = \text{softmax}\left( \frac{QK^T}{\sqrt{d_k}} \right) V \]

    Vision Transformers (ViTs) were introduced for image analysis in 2020 \cite{ViT}. Self-attention was calculated between an even sequence of small, non-overlapping patches. 
    Self-attention takes into account the importance of patches/pixels in relation to the elements of an image, making it highly adaptable. Convolutions only extract patch information locally, not globally.
  
    Transformers didn't start in vision. They were first built for text, where modeling broader context and long-range relationships is essential. The same self-attention that helped models like BERT and GPT capture meaning across sentences and paragraphs turned out to be just as powerful for images. This led to a shift where researchers started representing an image as a sequences of patches, making it possible to bring transformer strengths into vision tasks \cite{2025transformers}.

    Transformers look across the entire image at once and discover broader relationships that CNNs often miss. This becomes important in tasks where the larger context is important, like understanding the overall scene of a disaster. The same research \cite{2025transformers} shows that ViTs scale extremely well, outperforming CNNs when enough data is available. 

    These advantages motivated us to use transformers as a starting point for our work and to explore how different hyper-parameters and techniques influence model performance.

\section{Methods}

The dataset we used was the xBD challenge dataset \cite{xbd_paper}. In total there were 5598 in the train set and 1866 images in the test set, Each image contains a set of buildings (notated as ``polygons'') which can be classified into the following categories: [\textit{No damage}, \textit{Minor damage}, \textit{Major damage}, \textit{Destroyed}, \textit{unclassified}].

The dataset contains a skew towards polygons labeled as ``\textit{no damage}'' which are about 8-10 times more frequent than all of the damaged classes combined \cite{xbd_paper}. As shown in Table~\ref{tab:dataset_distribution}, the main xBD dataset exhibits a significant class imbalance, with ``\textit{no damage}'' representing the vast majority of the 425,368 total instances.

The distribution in the main dataset was mimicked in the challenge dataset. Aggregated from the confusion matrices for model evaluation, the distribution is approximated to be around 60.7\% no damage, 8.7\% minor damage, 14.2\% major damage, and 15.5\% destroyed.

\begin{table}[H]
\centering
\small 
\begin{sc} 
\begin{tabular}{lr}
\toprule
Class & Count \\
\midrule
No Damage & 313,033 \\
Minor Damage & 36,860 \\
Major Damage & 29,904 \\
Destroyed & 31,560 \\
Unclassified & 14,011 \\
\midrule
\textbf{Total} & \textbf{425,368} \\
\bottomrule
\end{tabular}
\end{sc}
\caption{Distribution of building damage labels in the dataset.}
\label{tab:dataset_distribution}
\end{table}

\begin{figure}[H]
\centering
\parbox{0.98\columnwidth}{\centering\textbf{no damage}}

\vspace{0.3em}

\begin{minipage}{0.48\columnwidth}
\centering
\includegraphics[width=\textwidth]{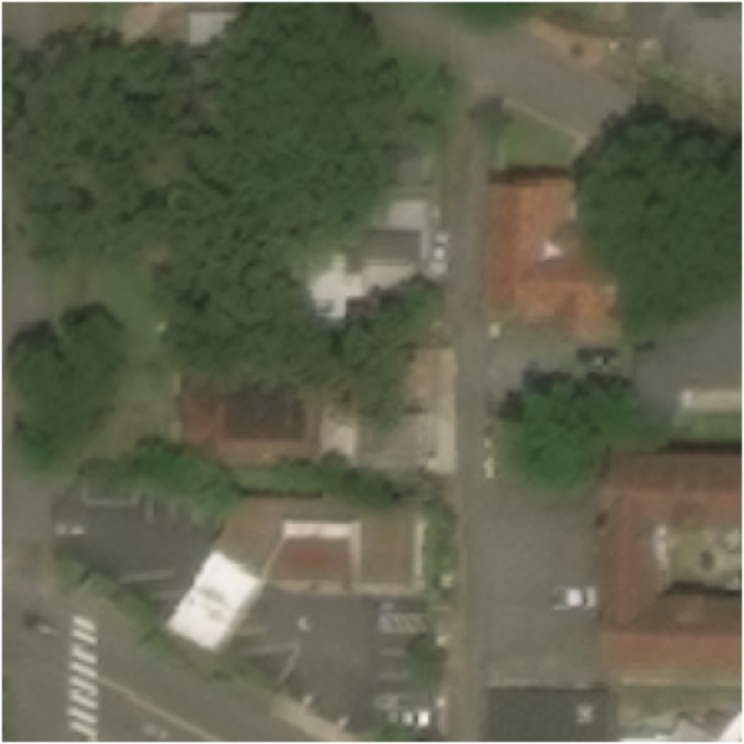}
\end{minipage}\hfill
\begin{minipage}{0.48\columnwidth}
\centering
\includegraphics[width=\textwidth]{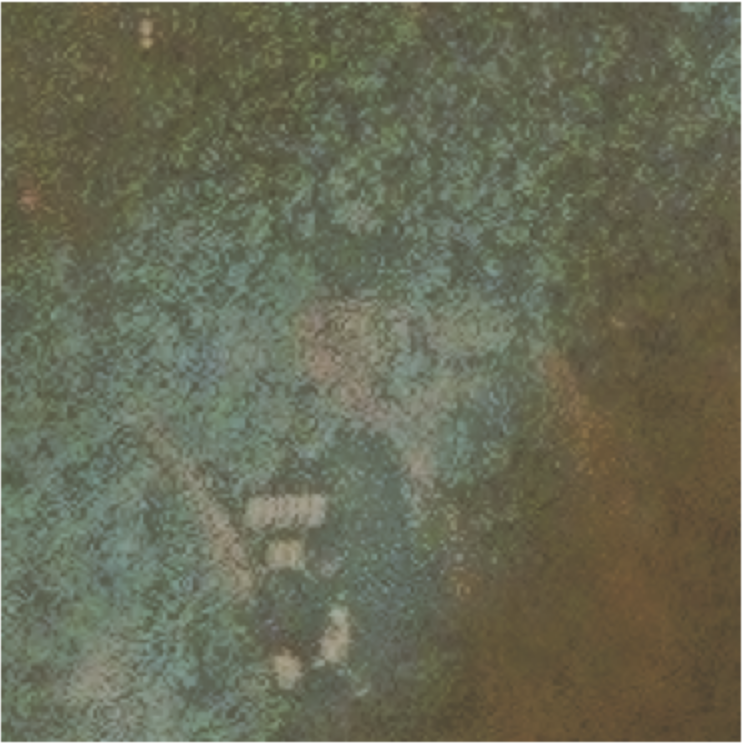}
\end{minipage}

\vspace*{0.5em}

\parbox{0.98\columnwidth}{\centering\textbf{minor damage}}

\vspace*{0.3em}

\begin{minipage}{0.48\columnwidth}
\centering
\includegraphics[width=\textwidth]{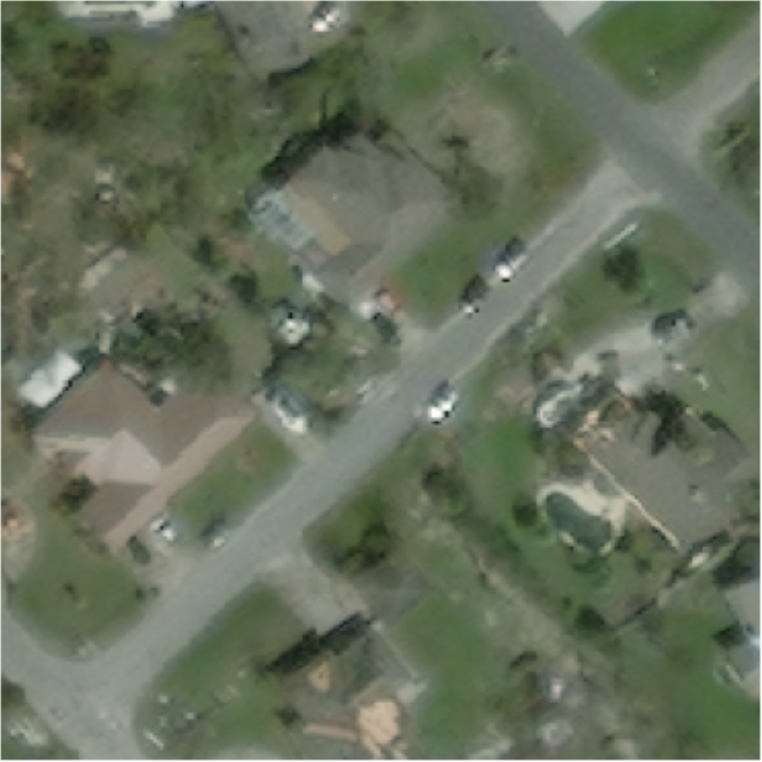}
\end{minipage}\hfill
\begin{minipage}{0.48\columnwidth}
\centering
\includegraphics[width=\textwidth]{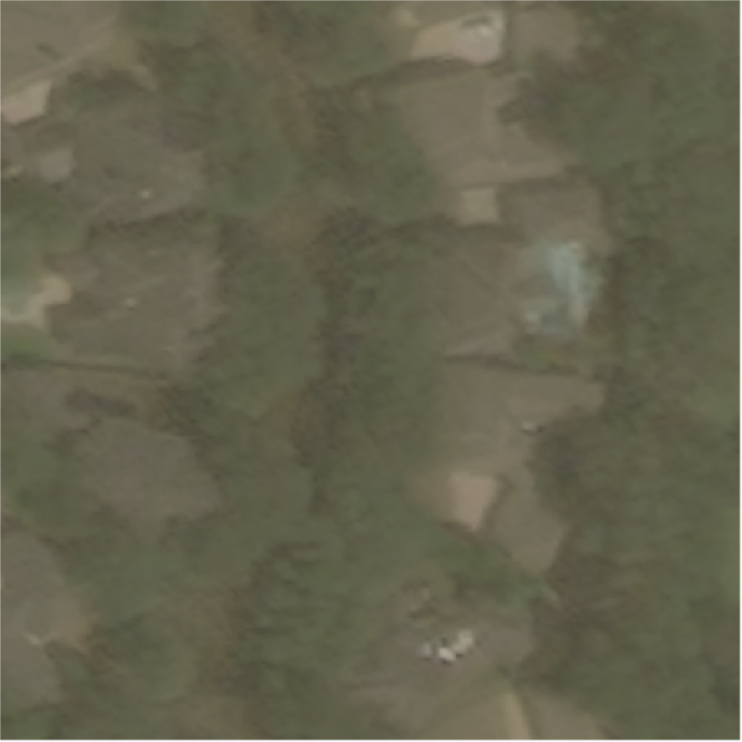}
\end{minipage}

\vspace*{0.5em}

\parbox{0.98\columnwidth}{\centering\textbf{major damage}}

\vspace*{0.3em}

\begin{minipage}{0.48\columnwidth}
\centering
\includegraphics[width=\textwidth]{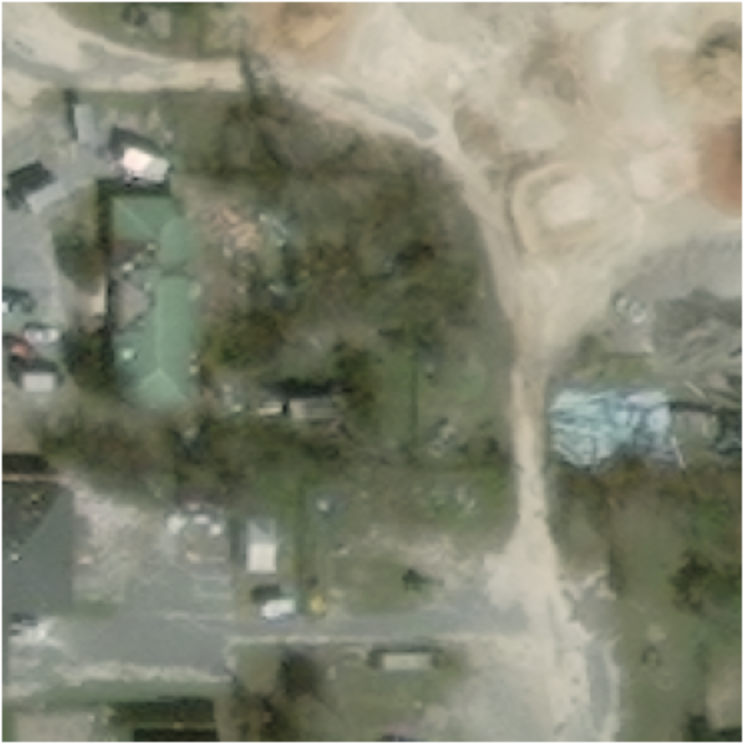}
\end{minipage}\hfill
\begin{minipage}{0.48\columnwidth}
\centering
\includegraphics[width=\textwidth]{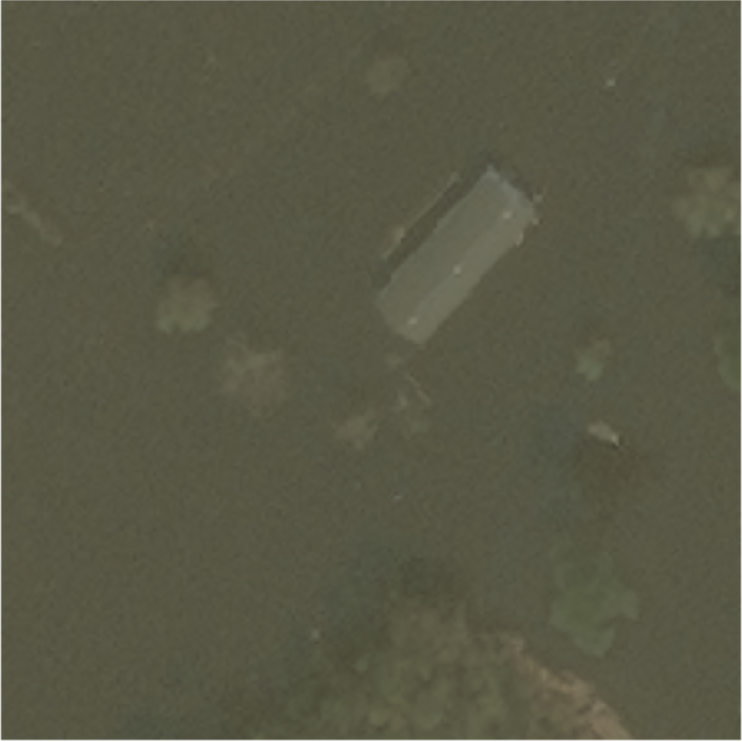}
\end{minipage}

\vspace{0.5em}

\parbox{0.98\columnwidth}{\centering\textbf{destroyed}}

\vspace{0.3em}

\begin{minipage}{0.48\columnwidth}
\centering
\includegraphics[width=\textwidth]{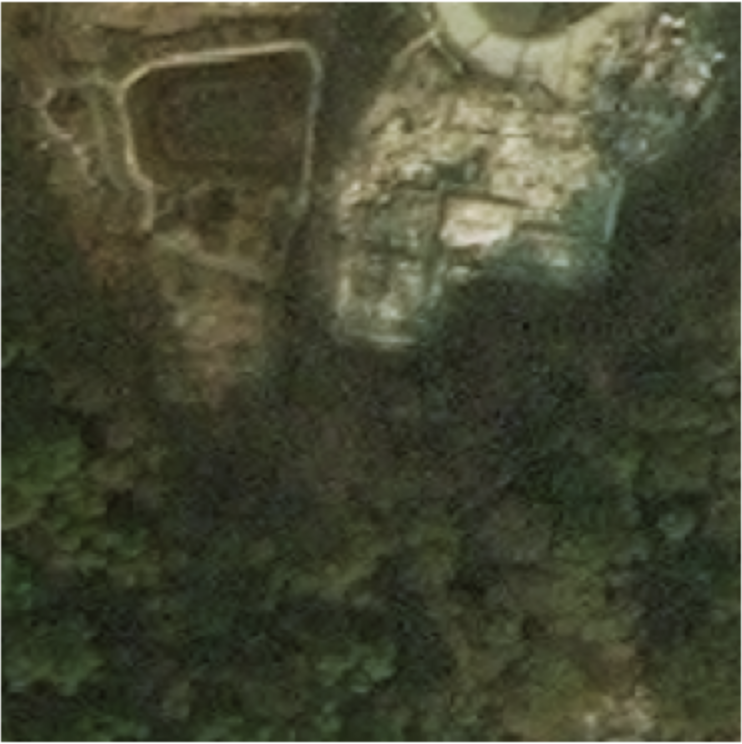}
\end{minipage}\hfill
\begin{minipage}{0.48\columnwidth}
\centering
\includegraphics[width=\textwidth]{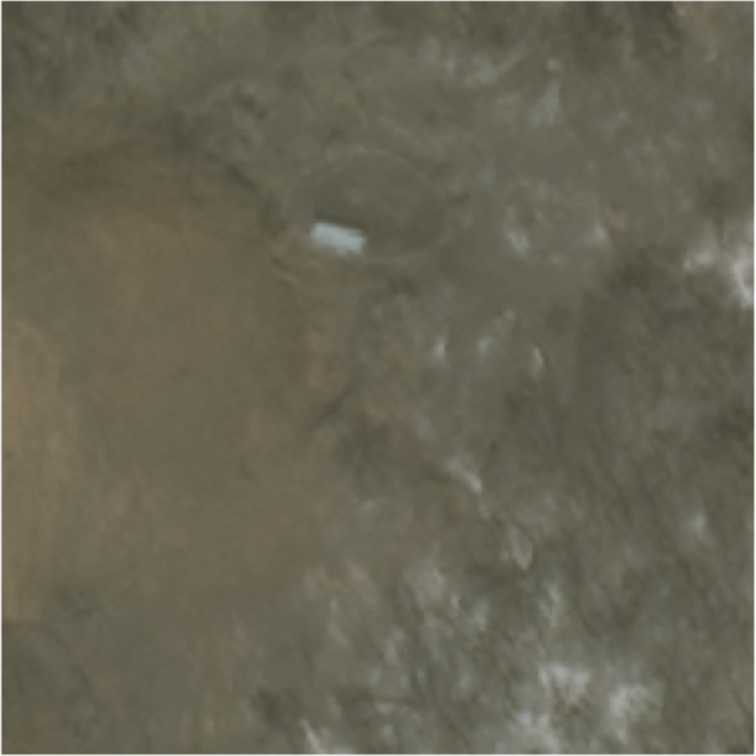}
\end{minipage}

\caption{Examples of image crops from each class in the xBD dataset. Each row contains two patches of the same class extracted from the training set dataloader.}
\label{fig:xbd-comparison1}
\end{figure}

 For the purpose of this project, unclassified images were ignored, focusing only on the four levels of damage classification. The dataset also includes masking options for each polygon. These additional features were ignored for the purpose of this project.

\subsection{Dataset and Preprocessing}

The custom dataloader filtered out images that do not have buildings with the subtypes in the label map: no-damage, minor-damage, major-damage, and destroyed. In this case, they return None instead of the image and label information. This removes around half of the images in the dataset. The remaining images were converted into NumPy arrays, to which a fully opaque (255) alpha channel was added to store transparency information along with the RGB data. 

From the valid buildings in the image, one was selected at random for patch extraction. A patch centered on the centroid coordinates of the building was cropped to fit the dimensions of the models (224x224 pixels for DeiT, 518x518 pixels for DINOv2). Images were cropped again with the polygons off-centered by a random factor within a search radius of 100 pixels. The ratio of transparent or black pixels was calculated for the new crop, entitled the ``empty ratio".

\vspace{1em}
\begin{minipage}{0.78\columnwidth}
\centering
\footnotesize
\begin{verbatim}
patch = img[y1:y1+self.patch_size, 
            x1:x1+self.patch_size, :]
rgb_patch = patch[:,:, :3]
alpha_patch = patch[:,:, 3]
black_pixels = np.all(rgb_patch <= 10, axis=2)
empty_pixels = np.logical_or(black_pixels, 
                           alpha_patch == 0)
empty_ratio = np.sum(empty_pixels) / 
              (self.patch_size**2)
\end{verbatim}
\label{fig:pixel_check}
\end{minipage}
\vspace{1em}

If the ratio of empty pixels exceeded 0.1 percent, the patch would have to be re-cropped again for a different random target within the search radius.


We calculated the fraction of fully transparent and black pixels in the crop and compared them to a threshold of 0.01. Images that exceeded this threshold were re-cropped to contain the fewest empty pixels within a defined window around the building's centroid.

The alpha channel was dropped before converting the NumPy array back into PIL format. Crops were passed through HuggingFace's image processor to be converted and normalized as a tensor compatible for the model. The custom dataset returned a dictionary containing the tensor and integer label information.

Datasets initialized with the custom dataset return class were unbatched. They were manually batched by passing them through HuggingFace's BatchIterableDataset, forming batches after filtering out any None values.
 
Figure~\ref{fig:xbd-comparison1} displays example image crops for each class.




\subsection{Compute Resource}

We ran our trials on Google Colab Pro+ in a Jupyter Notebook environment. This posed a few limitations. Most training trials were done with Colab Pro, which allocates 100 compute units. Our training took around 7 compute units per hour, and trials ranged from 4-9 hours, allowing for only 1-3 training sessions per month. We switched to a Pro+ subscription later on in our training process, which allocates 500 compute units per month.

Because Google Colab is cloud-based, training time varied depending on the number of users running on the notebook. Additionally, Pro only accounts for continuous runtimes up to 12 hours before cutting off. Pro+ allows runtimes up to 22 hours. It was observed that approximated runtimes consistently exceeded 24 hours when ran during the morning or afternoon. Training was mostly done overnight to avoid exceeding these thresholds. 

\subsection{Models}

We fine-tuned two models: the tiny Data-efficient Image Transformer [DeiT-Ti] and a smaller sized Distillation with No Labels, version two [DINOv2-small]. Both models utilize the ViT architecture, as well as the teacher-student model (knowledge distillation). The models were accessed via their HuggingFace implementation. 

\subsubsection{DeiT}

\begin{figure*}[h] 
\centering
\includegraphics[width=0.8\textwidth]{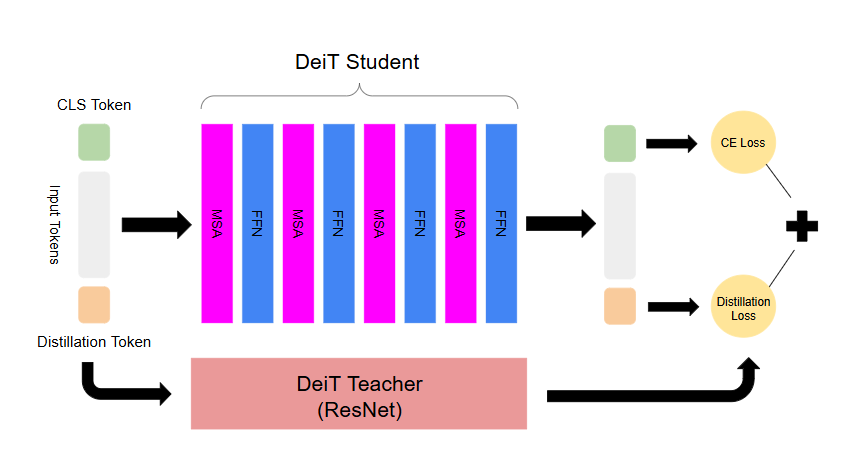} 
\caption{Structural diagram of the DeiT model. Based off Touvron 2021 \cite{DeiT}.}
\label{fig:deit_arch}
\end{figure*}

The DeiT based model we used was pre-trained trained on ImageNet \cite{imagenet} and has 5 million parameters \cite{DeiT}. Images (224x224) are divided into fixed patches (16x16) and flattened. They add an additional class [CLS] token for ground-truth classification as well as a distinctive distillation token \cite{DeiT}. 

The model has twelve layers, each containing three attention heads. For every multi-head self-attention layer [MSA] is a pair of linear layers  separated by a Gaussian Error Linear Unit [GELU], comprising the feed-forward network [FFN]. The MSA and FFN make up the transformer block. 

The CLS token goes through a linear layer to produce the classification logits for the ImageNet classes. The distillation token references model predictions from the teacher model (a ResNet), then produces it's own set of logits through a linear layer. The logits from the distillation token and CLS token are averaged for the final classification. 

The DeiT model is abstractly represented in Figure~\ref{fig:deit_arch}.

\subsubsection{DINOv2}

\begin{figure*}[h] 
\centering
\includegraphics[width=0.8\textwidth]{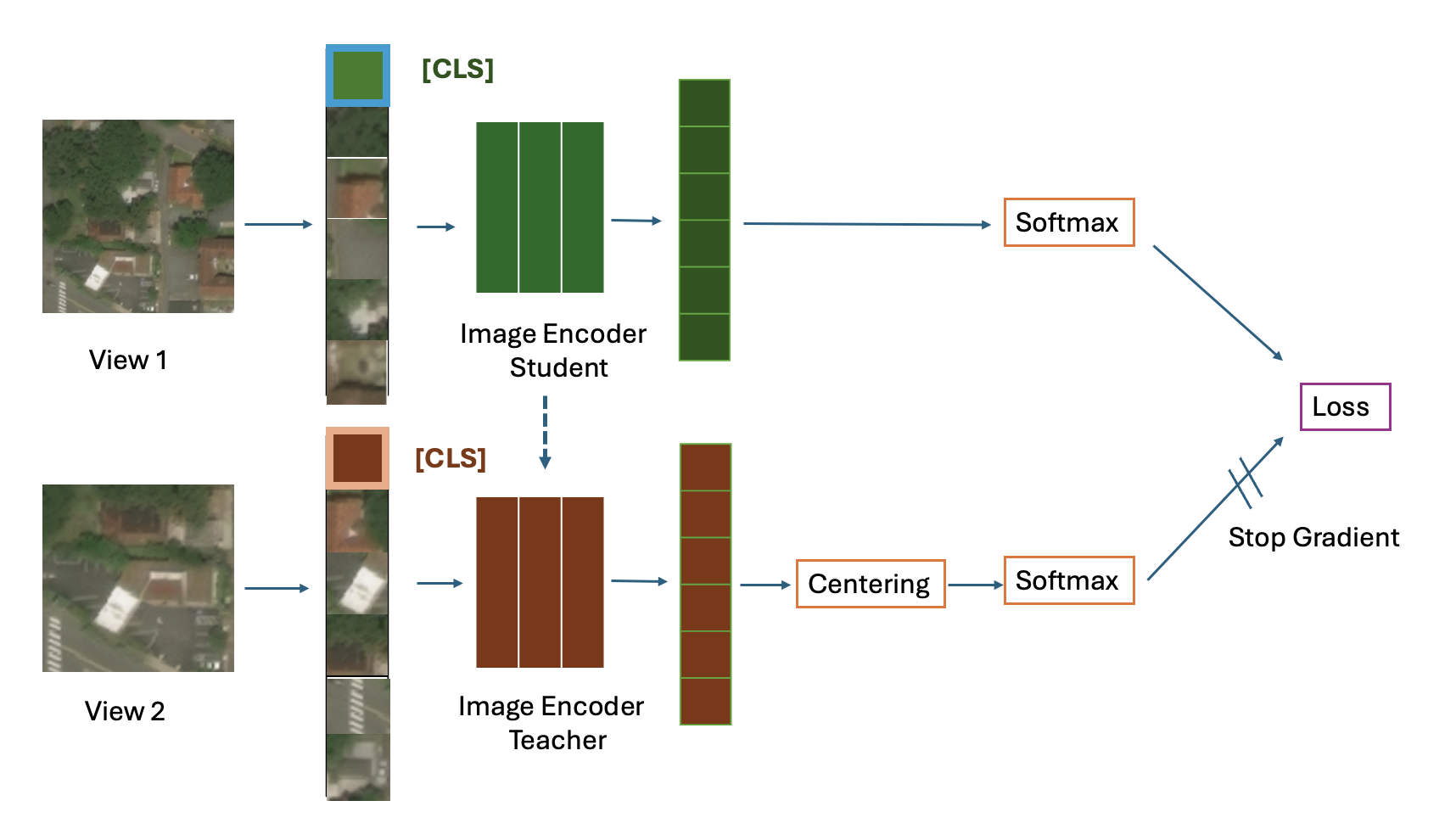} 
\caption{Structural diagram of the DINOv2 model. Based off Oquab 2023 \cite{DINOV2_paper}.}
\label{fig:dino_arch}
\end{figure*}

DINOv2 has 22 million parameters and was pretrained in a self-supervised manner on the LVD-142M dataset, containing 142 million images \cite{DINOV2_paper}. 

DINOv2 is comparable to the architecture for DeiT. The model also consists of 12 layers, but with 6 attention heads per layer rather than DeiT's 3 heads per layer \cite{DINOV2_paper}. The transformer block is identical to DeiT's (MSA followed by FFN), except it uses Swish-Gated Linear Units [SwiGLUs] instead of GELUs. Additionally the model uses a CLS token, but omits the distillation token used in DeiT.

Images are divided into fixed 14x14 patches and flattened with the additional CLS token for global representation in the logits \cite{DINOV2_paper}. After running through the layers, the model uses a self-distillation student-teacher network in contrast with the supervised student-teacher network in DeiT. The CLS token passes through a final linear to produce the classification logits. 

The DINOv2 model is abstractly represented in Figure~\ref{fig:dino_arch}.

\subsection{Training}

We trained on the A100 High-Ram GPU using Google Colab Pro+, which offered runtimes up to 24 hours long. Across all trials, training ranged from 4-9 hours. Run time was dependent on resource availability and user activity. 

Models were trained using a custom HuggingFace trainer, which overrides the compute loss to compute and log accuracy, precision, recall, and f1 scores on every training batch in real time.

\subsubsection{DeiT End-to-End}

We searched over the following learning rates in the grid search for the DeiT model:

\{1e-7, 1e-6, 1e-5, 5e-5, 1e-4\}

Each learning rate was tested in combination with the following batch sizes:

\{8, 16, 24, 32\}

Batch size was primarily constrained by compute; batch sizes of 32 or greater exceeded the 24 hours available to run on GPU.

Weight decay was set at 0.05. The maximum number of training steps was computed as:
\begin{equation}
\label{eq:steps}
\text{max\_steps} = \left\lceil  \frac{\text{dataset\_len} \times \text{num\_epochs}}{\text{train\_batch\_size} \times \text{num\_devices}} \right\rceil
\end{equation}%
Early training cycles ran for two epochs before narrowing down to the final models, ran at 5 epochs each.

The training set was split using an 80/20 random split for training and validation. The evaluation strategy was set every 20 training steps, which reported the accuracy, precision, and F1 score.
The evaluation batch size was set at 64 for all trials.

The training accuracy was evaluated with a running average window of 10 steps.

Checkpoints were saved every epoch.

\begin{figure}[H]
\centering
\includegraphics[width=\columnwidth]{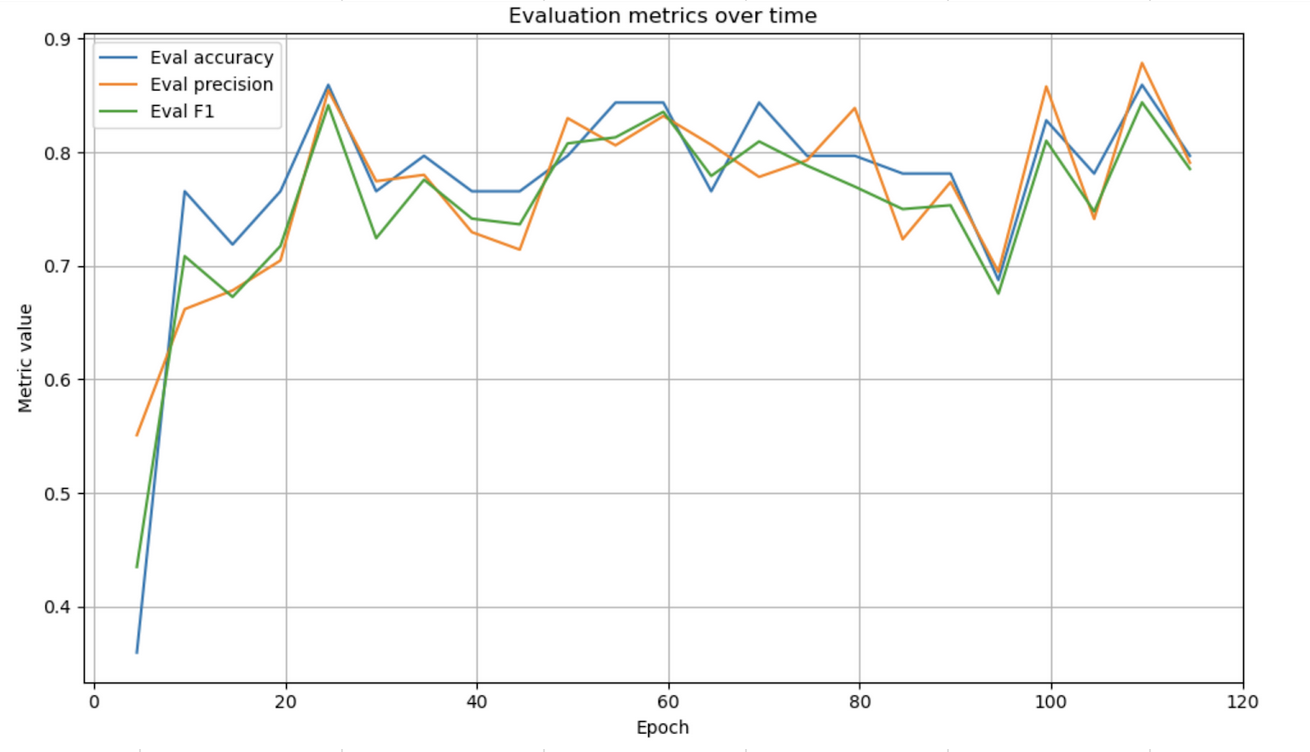}
\caption{Example plots of training accuracy and validation accuracy, precision and f1 for DeiT end-to-end model trained with learning rate 1e-5 and training batch size of 24 for 5 epochs. X-axis is scaled in step incrementations, y-axis with the metric scores.}
\label{fig:xbd-comparison}
\end{figure}

The highest performing model ran on 5 epochs was trained with a learning rate of 1e-5 and a training batch size of 24. The training curves for this model are shown in Figure~\ref{fig:xbd-comparison}.

\subsubsection{DINOv2 End-to-End}

We trained the DINOv2 model on the highest performing learning rates informed from the DeiT training: 1e-5 and 5e-5. Both models were trained on two epochs with training batch sizes of 8. The remaining hyperperameters were tuned exactly as they were for the DeiT model.

Given compute constraints, the training couldn't be evaluated beyond two epochs.

\subsubsection{Frozen Head Training}

Both the DINOv2 and DeiT models were trained with all weights frozen except for the classifier head. They ran for 10 epochs each, with a batch size of 24 and on a learning rate of 1e-3. The remaining hyperperameters remained constant from end-to-end training.




\section{Results}

\subsection{Evaluation}

Models were evaluated by pulling the weights from the checkpoint exhibiting the best validation and training accuracy performance.

The evaluation loop was tested on the test dataset, separate from the training and validation sets. The test dataset was preprocessed identically to the training and validation sets while creating the evaluation dataloader.

For each batch in the evaluation dataloader, the pixel values and ground truth labels were transferred to the CUDA device. Model forward passes generated the logits, and class predictions were derived via argmax over the class dimension. 

The full evaluation pass was repeated across 5 independent runs to increase the variety of buildings classified, mitigating the effects of the class imbalance on the evaluation results. 

The saved list of labels and predictions was used to calculate the weighted averages for accuracy, precision, recall, and f1 score. Confusion matrices displayed the classification performance across no damage, minor damage, major damage, and destroyed categories.

\begin{figure}
\centerline{\includegraphics[width=\columnwidth]{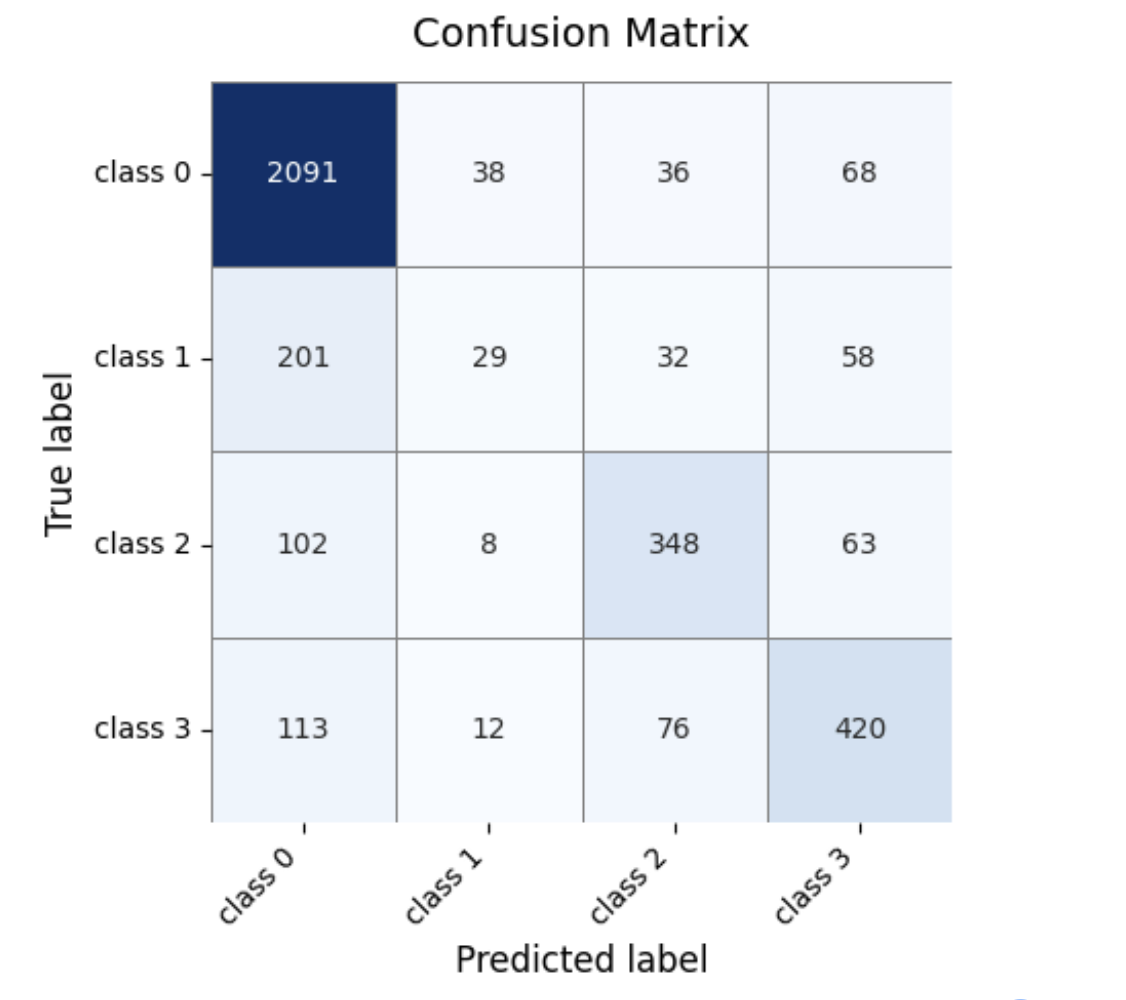}}
\caption{Confusion matrix for DeiT end-to-end model comparing predicted labels against ground truth for all four classes, where class 0 is ``no damage", class 1 is ``minor damage", class 2 is ``major damage" and class 3 is ``destroyed".}
\label{fig:confusion-matrices-comparison}
\end{figure}

The confusion matrix for the DeiT end-to-end model is represented in Figure~\ref{fig:confusion-matrices-comparison}.

\subsection{Results Table}

\begin{table*}[t] 
\centering
\begin{small}
\begin{sc}
\begin{tabular}{lccccc}
\toprule
Model & Accuracy & F1 Score & Precision & Recall \\
\midrule
DeIT (End-to-End) & 0.782 & 0.599 & 0.641 & 0.593\\
DeIT (Frozen) & 0.735 & 0.528 & 0.592 & 0.537\\
DINOv2 (End-to-End) & 0.760 & 0.565 & 0.620 & 0.579\\
DINOv2 (Frozen) & 0.736 & 0.526 & 0.553 & 0.534\\
ResNet (Macro Average) \cite{xbd_paper} & - & 0.320 & 0.576 & 0.271\\
Few-Shot (Macro Average) \cite{fewshot} & - & 0.638 & 0.643 & 0.633\\
UNet Encoder + ResNet18 \cite{chen2020} & - & 0.63 & 0.71 & 0.58\\
\bottomrule
\end{tabular}
\end{sc}
\end{small}
\caption{Comparison of different model architectures and training strategies.}
\label{tab:model_comparison}
\end{table*}


\vspace*{0.15in}

\begin{table*}[t] 
\centering
\begin{small}
\begin{sc}
\begin{tabular}{lcccc} 
\toprule
Class & F1 Score & Precision & Recall \\
\midrule
No Damage & 0.90 & 0.87 & 0.93\\
Minor Damage & 0.14 & 0.34 & 0.09\\
Major Damage & 0.73 & 0.76 & 0.71\\
Destroyed & 0.73 & 0.71 & 0.74\\
\bottomrule
\end{tabular}
\end{sc}
\end{small}
\caption{Performance across classes for DeiT end-to-end model.}
\label{sample-table}
\end{table*}

DeiT (E2E) achieved the highest overall performance, with an accuracy of 0.782 and an F1 score of 0.599, surpassing its frozen variant by over 7\% in accuracy and 0.07 in f1 score. Similarly, DINOv2 (E2E) performed better than its frozen counterpart, yielding an F1 score of 0.565 compared to 0.526 when frozen.

When benchmarked against prior works, both transformer-based models demonstrated notable improvements over traditional CNN baselines. The best-performing transformer model (DeiT E2E) exceeded the ResNet baseline \cite{DeiT}, which reported an F1 score of 0.320, and also matched or slightly outperformed the few-shot baseline \cite{fewshot}, which attained an F1 of 0.638. 

As compared to the UNet Encoder and ResNet method used by Chen et al., our macro-average performance was lower across accuracy, precision, recall, and F1. But the individual classification performance was superior to their performance for the No-damage (0.90 vs 0.88), major (0.73 vs 0.53), and destroyed classes (0.73 vs 0.55.

\section{Discussion and Conclusion}

In this project, we investigate whether vision transformers could handle messy, imbalanced data, as seen in the xBD dataset. We started by building a patch-based preprocessing pipeline that strips out most of the background (such as clouds, roads etc.) and ``zooms in" on the buildings. The idea was to increase the amount of useful signals that we have to train our models.

The next step was to fine-tune the DeiT and DINOv2 models. Even using limited Colab resources, our model consistently outperformed earlier CNN, few shot, ResNet and other earlier benchmarks, especially for the severe damage classes. Our model struggled with the minor damage class where the visual cues were more subtle and hard to distinguish. 

Looking ahead, this work leads towards a few natural next steps. 
(1) On the data side, we can be more intentional on which buildings we sample and explicitly address class imbalance. For e.g., we could reduce the number of ``No-damage" class buildings. We could also use Generative AI simulation techniques to create more samples for the other classes. This way the model does not spend most of its capacity on ``No-damage" cases by default.
(2) On the modeling side, it would be useful to move beyond single crops of building images and consider multiple buildings based on proximity, simultaneously. This way the model can pick up on the broader scene and overall impact of damage in the area.
(3) Finally, we can simplify the label space for the images into three buckets - no damage, medium severity damage, and high-severity damage. This makes it better aligned to how humans actually triage and respond to disaster. Additionally, this might alleviate the problem with the subtle difference between minor damage and no damage classes.

Together the above steps could push transformer based models as a strong benchmark for reliable disaster classification models.

\section*{Software and Data}

The xBD dataset is available for download on the xView2 challenge website: \url{https://xview2.org/dataset}

The Colab notebook used to run this project is available on GitHub: \url{https://github.com/smritsiv27/Classifying-Satellite-Images-with-Vision-Transformers}.

\bibliography{example_paper}
\bibliographystyle{icml2025}





\end{document}